\documentclass[reprint,aps,pre]{revtex4-1}

\usepackage{amsmath,amsfonts,amscd,amssymb,graphicx,subeqnar}
\usepackage[table,xcdraw]{xcolor}

\DeclareMathOperator*{\argmin}{arg\rm{}min}

\usepackage{dcolumn}
\usepackage{multirow}
\usepackage{bm}
\usepackage[percent]{overpic}
\usepackage{color}
\usepackage{mathrsfs}
\usepackage{adjustbox}
\usepackage{rotating}
\usepackage{float}
\makeatletter
\let\newfloat\newfloat@ltx
\makeatother
\usepackage{algorithm}
\usepackage{algpseudocode}
\algnewcommand\algorithmicinput{\textbf{Input:}}
\algnewcommand\Input{\item[\algorithmicinput]}
\usepackage{soul}
\usepackage{stackengine}

\usepackage[hypertexnames=false,
            colorlinks=true,
            linkcolor=blue,
            citecolor=blue,
            urlcolor=blue]
            {hyperref} 

\usepackage{lipsum}  

\begin{document}

\title{Leveraging arbitrary mobile sensor trajectories with shallow recurrent decoder networks for full-state reconstruction}

\author{Megan R. Ebers$^{\dag}$, Jan P. Williams$^{\dag}$, Katherine M. Steele$^{\dag}$, J. Nathan Kutz$^{*, **}$}
   \affiliation{$^\dag$Department of Mechanical Engineering, University of Washington, Seattle, WA 98195}
 \affiliation{$^*$Department of Applied Mathematics, University of Washington, Seattle, WA 98195} %
 \affiliation{$^{**}$Department of Electrical and Computer Engineering, University of Washington, Seattle, WA 98195}

\begin{abstract}
Sensing is one of the most fundamental tasks for the monitoring, forecasting and control of complex, spatio-temporal systems.  
In many applications, a limited number of sensors are mobile and move with the dynamics, with examples including wearable technology, ocean monitoring buoys, and weather balloons.
In these dynamic systems (without regions of statistical-independence), the measurement time history encodes a significant amount of information that can be extracted for critical tasks. 
Most model-free sensing paradigms aim to map current sparse sensor measurements to the high-dimensional state space, ignoring the time-history all together. 
Using modern deep learning architectures, we show that a sequence-to-vector model, such as an LSTM (long, short-term memory) network, with a decoder network, dynamic trajectory information can be mapped to full state-space estimates.
Indeed, we demonstrate that by leveraging mobile sensor trajectories with shallow recurrent decoder networks, we can train the network (i) to accurately reconstruct the full state space using arbitrary dynamical trajectories of the sensors, (ii) the architecture reduces the variance of the mean-square error of the reconstruction error in comparison with immobile sensors, and (iii) the architecture also allows for rapid generalization (parameterization of dynamics) for data outside the training set.  
Moreover, the path of the sensor can be chosen arbitrarily, provided training data for the spatial trajectory of the sensor is available. 
The exceptional performance of the network architecture is demonstrated on three applications: turbulent flows, global sea-surface temperature data, and human movement biomechanics.
\end{abstract}

\maketitle

\section{Introduction}

\begin{figure*}[t]
    \centering
    \includegraphics[width=\textwidth]{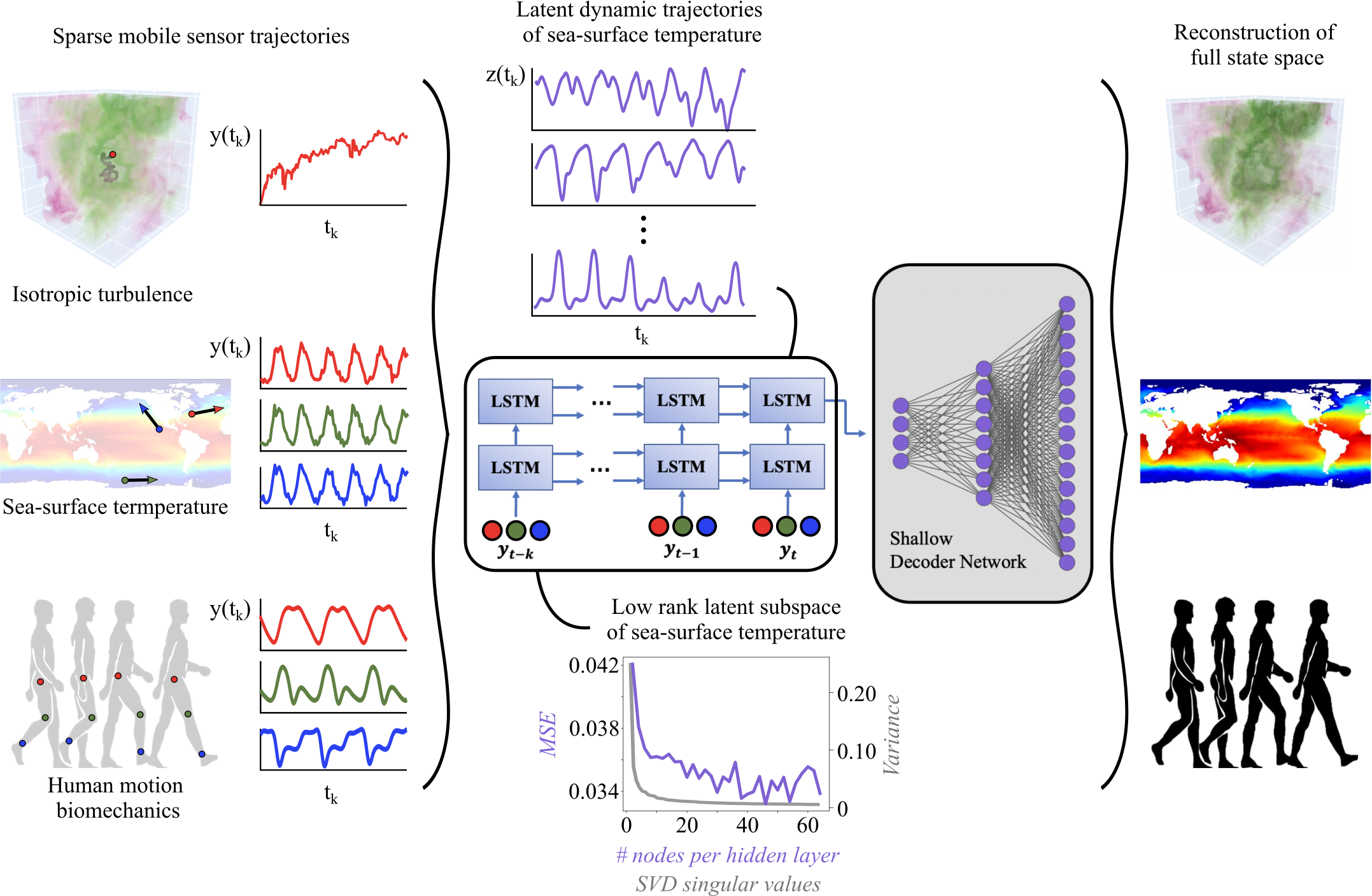}
    \caption{Summary figure of a \textit{shallow recurrent decoder network} (SHRED) leveraging mobile sensors to reconstruct full state-space estimates from sparse dynamical trajectories. (Left) Sensor trajectory history encodes global information of the spatio-temporal dynamics of the sparsely measured system. In this work, we evaluate three challenging datasets, including forced isotropic turbulence, global sea-surface temperature, and human biomechanics. (Middle) The mobile SHRED architecture can (i) embed the multiscale physics of a system into a compact and low-dimensional latent space, and (ii) provide a mapping from the sparse mobile sensors to a full state estimate. (Right) The high-dimensional and complex system states can be reconstructed, provided training data for the dynamical trajectory of the sensor(s) is available.}
    \label{fig:Figure1}
\end{figure*}

Sensing is a ubiquitous task, being critically important to every scientific and engineering discipline~\cite{manohar2018data}.  Sensor technologies accelerate scientific advancement by providing improved observational data that is capable of advancing new scientific theories and improving technological development.  From the James Webb telescope~\cite{gardner2006james} to single neuron recordings~\cite{stevenson2011advances}, sensors provide an interface with the physical world that allows us to interrogate diverse and complex systems across scientific domains.  In many application fields, sensors are mobile by design, moving with dynamical trajectories as they record quantities of interest.  Application areas such as motion tracking of human biomechanics or robotic systems, buoys that measure ocean dynamics, and weather balloons are all examples of technologies where sensors are not (or cannot be) immobile. From these mobile sensors, some or all of the canonical tasks of sensing are enacted, including reconstruction, forecasting, model discovery, control and uncertainty quantification~\cite{manohar2018data}.  Emerging sensor technologies are capable of producing exceptional quantities of data which can now be leveraged by machine learning algorithms.  Specifically, we utilize a recurrent neural network~\cite{medsker2001recurrent,salehinejad2017recent,zaremba2014recurrent} to learn the temporal sequences in a latent space that maps to full state-space estimates via a shallow decoder network~\cite{erichson2020shallow,williams2022data}.  This {\em shallow recurrent decoder network} (SHRED) architecture has been successful for immobile sensors~\cite{williams2023sensing}, but this work shows how it can be used in a mobile sensing framework.  Indeed, mobile sensing affords distinct advantages over immobile sensors, including the reduction of the variance of the mean-square reconstruction error in comparison with immobile sensors and improved generalization (parameterization of dynamics) for data outside the training set.  

Traditional system identification techniques such as the Kalman filter and its variants have long achieved state-of-the-art performance \cite{law_data-2015}, but are limited to applications where a model of the underlying dynamics is readily available.  With the recent explosion in quantity and quality of data from high-dimensional systems, we are increasingly required to perform state estimates of systems with unknown or computationally infeasible dynamics.  State estimation in these cases is typically achieved by exploiting low-rank structures in the data learned by a singular value decomposition (SVD), otherwise known as proper orthogonal decomposition (POD) \cite{kutz2013data, brunton2019data}. Gappy POD was one of the first such techniques to reconstruct high-dimensional fields from limited, sparse sensor measurements \cite{everson_karhunenloeve_1995}.  Since then, much work has been aimed at optimizing sensor placement to well-condition the linear inverse problem upon which state estimates rely \cite{manohar2018data, drmac_new_2016, barrault_empirical_2004, chaturantabut_nonlinear_2010}. There exist numerous techniques for determining optimal sensor locations in small search spaces \cite{boyd2004convex, joshi_sensor_2009, caselton_optimal_1984, krause_near-optimal_2008, lindley_measure_1956, sebastiani_maximum_2000, paninski_asymptotic_2005}, but in high-dimensional systems a greedy approximation, such as QR decomposition with column pivoting, is needed.  These techniques for state-estimation in the absence of a governing model also differ from Kalman filtering in that they only utilize static measurements, rather than condition upon a history of sensor data.  While this lends the methods flexibility to reconstruct data from a dataset of faces or a dynamical system, it ignores the rich information that can be gleaned from incorporating past sensor measurements in a method for reconstruction. 



Leveraging the temporal histories of sensor measurements offers an opportunity to capitalize on both the rich information from measurement trajectories and enable greater inclusion of measurement modalities -- beyond immobile sensors. To date, the inclusion and design of dynamic trajectories are addressed with sensor path planning. Path planning is a long-standing challenge in the field of engineering and robotics related to navigation and estimation in dynamical environments \cite{gunnarson2021learning, krishna2022finite, buzzicotti2020optimal, biferale2019zermelo, madridano2021trajectory}. 
The majority of path planning has focused on the modeling and control of the mobile sensor positions \cite{shriwastav2022dynamic, lynch2008decentralized, leonard2007collective, devries2013observability, ogren2004cooperative, zhang2010cooperative, paley2020mobile, peng2014dynamic}.
For example, recent work proposed the use of mobile sensors with Kalman filtering to estimate spatio-temporal data \cite{mei2022mobile}. The authors' approach to sensor path planning for dynamic estimation includes two main steps: (i) dynamic mode decomposition finds a low-rank representation of the data for Kalman filtering, and (ii) greedy path finding optimizes the observability matrix along the path and improves Kalman filter estimation.
However, path planning for mobile sensing and state estimation may be unnecessary. 
The SHRED architecture has demonstrated the successful reconstruction of complex spatio-temporal data with arbitrarily placed -- albeit immobile -- sensors \cite{williams2023sensing}. Therefore, this work aims to employ mobile sensors with SHRED to understand whether the use of arbitrary dynamic trajectories can feasibly and reliably reconstruct complex spatio-temporal data, thereby removing the challenge of path planning altogether.

As a mathematical architecture, SHRED leverages emerging deep learning paradigms \cite{lecun2015deep} and their universal approximation capabilities \cite{hornik1989multilayer} to map from sparse and minimal sensor measurements to high-dimensional spatio-temporal data.  In contrast with common sensing modalities where principled sensor placement is critical~\cite{manohar2018data}, the LSTM embedded in the SHRED architecture extracts a latent representation of the low-dimensional dynamics using sensor trajectory (time history) information which is agnostic to the sensor location.  Thus the sensor trajectory history encodes {\em global} information of the spatio-temporal dynamics of the measured system.  And while previous research has considered mobile sensor features like the timescale of the spatio-temporal dynamics, velocity of the sensors, and rate of sampling as part of the path planning optimization \cite{mei2022mobile}, in this work, such features are treated hyperparameters to be tuned in the SHRED architecture.  We demonstrate the performance of SHRED on three challenge data sets related to turbulence, sea-surface temperature, and human biomechanics. In all three cases, the mobile SHRED architecture provides a high-quality and accurate algorithm for estimating the original complex, multiscale, and high-dimensional system. In all the examples, it is demonstrated that by leveraging mobile sensor trajectories with SHRED, we can train the network to accurately reconstruct the full state space while also reducing the variance of the mean-square error of the reconstruction error in comparison with immobile sensors.  Moreover, the path of the sensor can be chosen arbitrarily, provided training data for the spatial trajectory of the sensor is available. 
SHRED works with high-probability given that the data do not have regions of statistically independent data.  In addition, when full state data is not available, proxy computational data with empirically similar data can be used for training.

\section{Mathematical Formulation}



The shallow recurrent decoder network (SHRED) can be understood as the amalgamation of an LSTM for processing a time-series of sensor measurements followed by a feedforward neural network, or decoder, for reconstructing a high-dimensional state from the learned latent representation of the LSTM \cite{williams2023sensing}. Let the high-dimensional state to be reconstructed be denoted as $x_T \in \mathbb R ^n$ and assume access to a set of sensor measurements $y_t = C x_t \in \mathbb R ^m$ for $t \in \{T - K, T - K +1, \dots, T -1, T \}.$  $K$ can be determined by empirical analyses of the system at hand.  We assume the measurements are sparse point measurements, that is $ m << n$ and $C$ consists of rows of the $n \times n$ identity matrix, although there is some evidence to suggest that neural network based reconstructions can be performed with nonlinear measurements \cite{erichson2020shallow}.  The set of sensor measurements serve as inputs to an LSTM \cite{hochreiter_long_1997} with recursive update equations
\begin{gather}
    h _t = \sigma \left( W _o \begin{bmatrix}  h_{t-1}, y_t \end{bmatrix} + b _o \right) \odot \tanh(c _t)\\
    c_t = \sigma \left( W _f \begin{bmatrix}
        h _{t-1}, y_t  
    \end{bmatrix} +b _f \right) \odot c _{t-1} \\ 
    + \sigma \left( W _i \begin{bmatrix}
        h _{t-1}, y_t
    \end{bmatrix} +b _f \right) \odot \tanh \left( W_g \begin{bmatrix}
        h_{t-1}, y_{t}
    \end{bmatrix} + b _g \right) \notag
\end{gather}
where $W _{RN} = \{W _o, W _f, W _i, W _g, b _o, b _f, b _i, b _g \}$ are the trainable weights and biases of the LSTM. We denote
\begin{equation}
    h_T = \mathcal G ( \{ y_t \} _{T - K} ^{T}; W_{RN}).
\end{equation}
The latent state $h_T$ learned by the LSTM has a variety of interesting properties that will be discussed in a later section.  

The feedforward component of the SHRED architecture is a shallow decoder with $b$ layers denoted by
\begin{equation}
    \mathcal F (h; W_{SD} ):= R(W^bR(W^{b-1} \cdots R (W^1h))),
\end{equation}
parameterized by trainable weights $W_{SD} = \{ W^1, \dots, W^b \}$ and with nonlinear scalar activation function $R$ (chosen to be ReLU). In total, the SHRED network is given by 
\begin{equation}
    \mathcal H ( \{ y_t \} _{T - K} ^{T}) =  \mathcal F (\mathcal G ( \{ y_t \} _{T - K} ^{T}; W_{RN}); W_{SD}).
\end{equation}
The network is trained to minimize reconstruction loss over a set of training states $\{x_t \}_{1}^{N}$,
\begin{equation}
    \mathcal{H} \in \argmin_{\widetilde{\mathcal{H}} \in  \mathscr{H}}  \sum _{t=1}^N ||x_i - \widetilde{\mathcal{H}}\left( \{ y _i \}_{i=t-K}^t \right)||_2,
\end{equation}
using the ADAM optimizer \cite{kingma_adam_2017}.  The assumption of access to a set of high-dimensional states for training in this manner is a strong one; simultaneous measurement of an entire high-dimensional system is sometimes simply impossible.  In such cases, a high-fidelity simulation can be used to train the network, provided the simulation accurately approximates the statistics of the real system.  Alternatively, if full-state measurements are possible, but prohibitively expensive in the long-term, the generation of training data can be viewed as a one-time up front cost.  

Previous work has demonstrated that such networks outperform traditional, POD based techniques for state estimation while requiring fewer available sensors \cite{williams2023sensing}.  In this work, we consider the case of a time-dependent measurement matrix $C.$  That is, 
\begin{equation}
    y_t = C_t x_t
\end{equation}
where $C_t$ varies in time.  This corresponds to a mobile sensor, in contrast to existing work which considers only immobile sensors.  We emphasize that while the set of measurement matrices $C_t$ can be chosen arbitrarily, corresponding training data is necessary in order to train the network; the architecture as it stands cannot extrapolate to unseen sensor location trajectories.  In the cases of periodic or quasi-periodic phenomena, this requirement amounts to the evolution of $C_t$ being periodic over some characteristic time-scale in order to extrapolate beyond a temporally partitioned training dataset. 

Despite the reputation of neural networks as lacking interpretability, we believe that there are a variety of potential theoretical connections between the latent state learned by the LSTM of a SHRED model and other domains of mathematics. In particular, there are parallels between the function of the LSTM and Takens’ embedding, which states that with a sufficiently long time-history, the dynamics of a time-delayed state variable are diffeomorphic to the dynamics of the high-dimensional state space. While the LSTM does not explicitly construct a time-delayed state, we conjecture that, once trained, it does so implicitly in its processing of sensor measurements. The feedforward network that follows the LSTM layer might then be thought of as having learned a diffeomorphism between the analog of a time-delayed vector and the high-dimensional state space. 

We also believe that theoretical exploration of the the LSTM's latent dimension size is warranted.  The plot in Fig. \ref{fig:Figure1} show the MSE of reconstructions of sea-surface temperature vs the number of nodes per hidden layer and the SVD singular value spectrum. Both plots exhibit an clear elbow at $\sim 5$ nodes per hidden layer and the $\sim 5$ singular value.  In traditional dimensionality reduction terms, the data exhibits a low-rank structure and can be well approximated using the $\sim 5$ dominant modes learned by the SVD.  We conjecture that a similar relation might exist between the minimum required LSTM state dimension and the underlying structure of the data.  In that case, the values of each entry in the latent state might be understood in analogy to the entries in the matrix $V$ of the SVD $X = U \Sigma V^T.$ The plot in Fig. \ref{fig:Figure1} illustrates that these entries do appear to evolve as some quasi-periodic phenomena. 

\begin{figure*}[t]
    \centering
    \includegraphics[width=\textwidth]{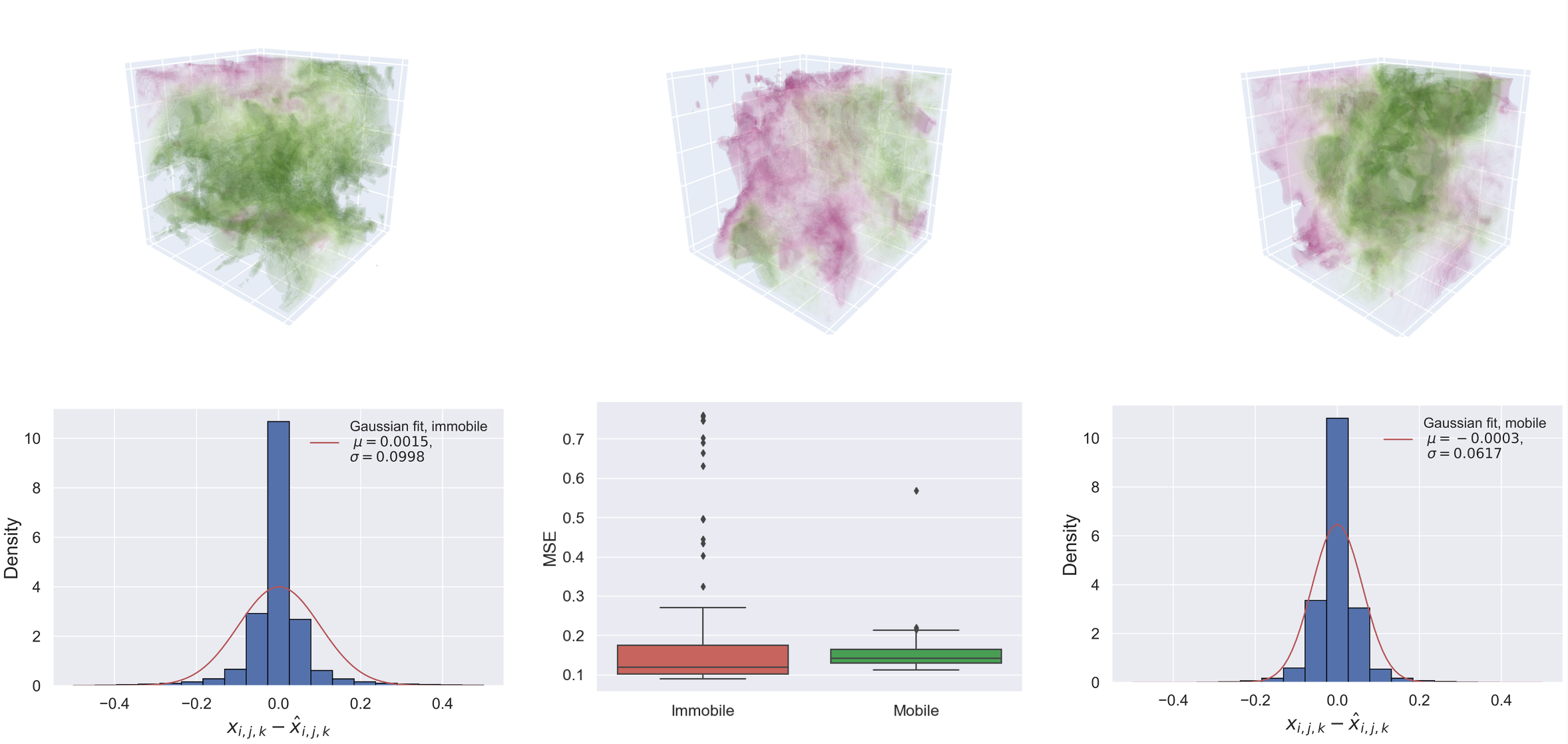}
    \caption{(Top) Example flow fields to be reconstructed from the test set. (Bottom) Histograms of the difference between ground truth and reconstruction across all nodes and samples in the test set for immobile sensors (left) and mobile sensors (right). Both distributions are approximately zero mean, but the variance for mobile sensors is lower. (Center) Box plot of the MSE evaluated across all samples in the test set for the 100 SHRED models with mobile and immobile sensors. While the median performance is similar, there are notably more outliers with poor performance in the case of immobile sensors.}

    \label{fig:TurbFig}
\end{figure*}

\section{Mobile sensing applications}

To demonstrate the mobile SHRED architecture, three example problems are selected:  isotropic turbulence, sea-surface temperature and human biomechanics.  The physics of all three systems produce high-dimensional, multiscale dynamics which is challenging to model in practice.  However, we will show in the figures that follow that SHRED provides exceptional performance by mapping between arbitrary trajectories of sensor measurements and the full state space.

\subsection{Forced isotropic turbulent flow}
The first dataset upon which we evaluate the performance of SHRED with mobile sensors is a forced isotropic turbulent flow from the Johns Hopkins turbulence database \cite{li_public_2008}.  The data was generated via the pseudo-spectral method with $1024^3$ nodes and utilized time steps of 0.0002 seconds.  From the simulation, we select a $50 \times 50 \times 50$ node volume with stride length of 4 in all dimensions. We select 1,257 snapshots of this volume with temporal spacing of 0.024 seconds. Of these 1,257 snapshots, the first 100 are set aside to account for our use of a time-history of 100 sensor measurements, 900 of the remaining snapshots are then randomly selected for training while the remainder are divided evenly between validation and test sets.  In this example, we do not consider a strict, temporal partition of training and test data. 

We train 100 SHRED models with one immobile, randomly placed sensors and 100 SHRED models with one mobile sensor following a modified random walk.  Each random walk is initialized from the center of the considered volume and every three time-steps takes a one-node step in one of the three spatial dimensions, each with equal probability.  With high probability, SHRED models with either mobile or immobile sensors accurately reconstruct the high-dimensional state as measured by MSE across the test set.  However, the box plot in Fig. \ref{fig:TurbFig} shows that a greater proportion of models using mobile sensors achieve good performance in comparison to the models using immobile sensors. The histograms in Fig. \ref{fig:TurbFig} plot the deviation of the reconstruction from the ground truth at each node and across all samples in the test set.  The estimates for both mobile and immobile sensors appear to be unbiased ($\mu = 0.0015$ immobile, $\mu = -0.0003$ mobile), but the variance of the mobile sensing reconstructions is less ($\sigma ^2 = 0.0998^2$ immobile, $\sigma ^2 = 0.0617^2$ mobile). Neither distribution is well approximated by a Gaussian, although the fit for mobile sensors is better.

\begin{table*}[t]
\def\arraystretch{1.5}
\resizebox{0.6\textwidth}{!}{
\begin{tabular}{|ccc|}
\hline
\multicolumn{3}{|l|}{Sea surface temperature reconstruction error}                              \\ \hline
\multicolumn{1}{|c|}{Route} &
  \multicolumn{1}{c|}{\begin{tabular}[c]{@{}c@{}}MSE \\ (random parition)\end{tabular}} &
  \begin{tabular}[c]{@{}c@{}}MSE\\ (temporal partition)\end{tabular} \\ \hline
\multicolumn{1}{|c|}{Immobile sensors (3)}    & \multicolumn{1}{c|}{0.0279} & 0.0460 \\ \hline
\multicolumn{1}{|c|}{Immobile sensor (1)}     & \multicolumn{1}{c|}{0.0317} & 0.0481 \\ \hline
\multicolumn{1}{|c|}{Atlantic Ocean}                  & \multicolumn{1}{c|}{0.0358} & 0.0442 \\ \hline
\multicolumn{1}{|c|}{Antarctica}                      & \multicolumn{1}{c|}{0.0321} & 0.0516 \\ \hline
\multicolumn{1}{|c|}{USA West Coast}                  & \multicolumn{1}{c|}{0.0321} & 0.0499 \\ \hline
\multicolumn{1}{|c|}{Atlantic Ocean + Antarctica}     & \multicolumn{1}{c|}{0.0395} & 0.1549 \\ \hline
\multicolumn{1}{|c|}{Atlantic Ocean + USA West Coast} & \multicolumn{1}{c|}{0.0440} & 0.1349 \\ \hline
\multicolumn{1}{|c|}{Antarctica + USA West Coast}     & \multicolumn{1}{c|}{0.0328} & 0.1278 \\ \hline
\multicolumn{1}{|c|}{\begin{tabular}[c]{@{}c@{}}Atlantic Ocean + Antarctica + \\ USA West Coast\end{tabular}} &
  \multicolumn{1}{c|}{0.0342} & 0.1345 \\ \hline
\end{tabular}%
}
\caption{Mean-squared error for SHRED reconstructing sea-surface temperature for both randomly and temporaly partitioned training/test/validation data. Dynamic trajectories from mobile sensors in the Atlantic Ocean, Antarctic Ocean, and along the USA's West Coast were compared to one and three immobile sensors for reconstruction the complex spatio-temporal sea-surface temperature data \cite{reynolds2002improved}.}
\label{tab:SST}
\end{table*}

\begin{figure*}[t]
    \centering
    \includegraphics[width=\textwidth]{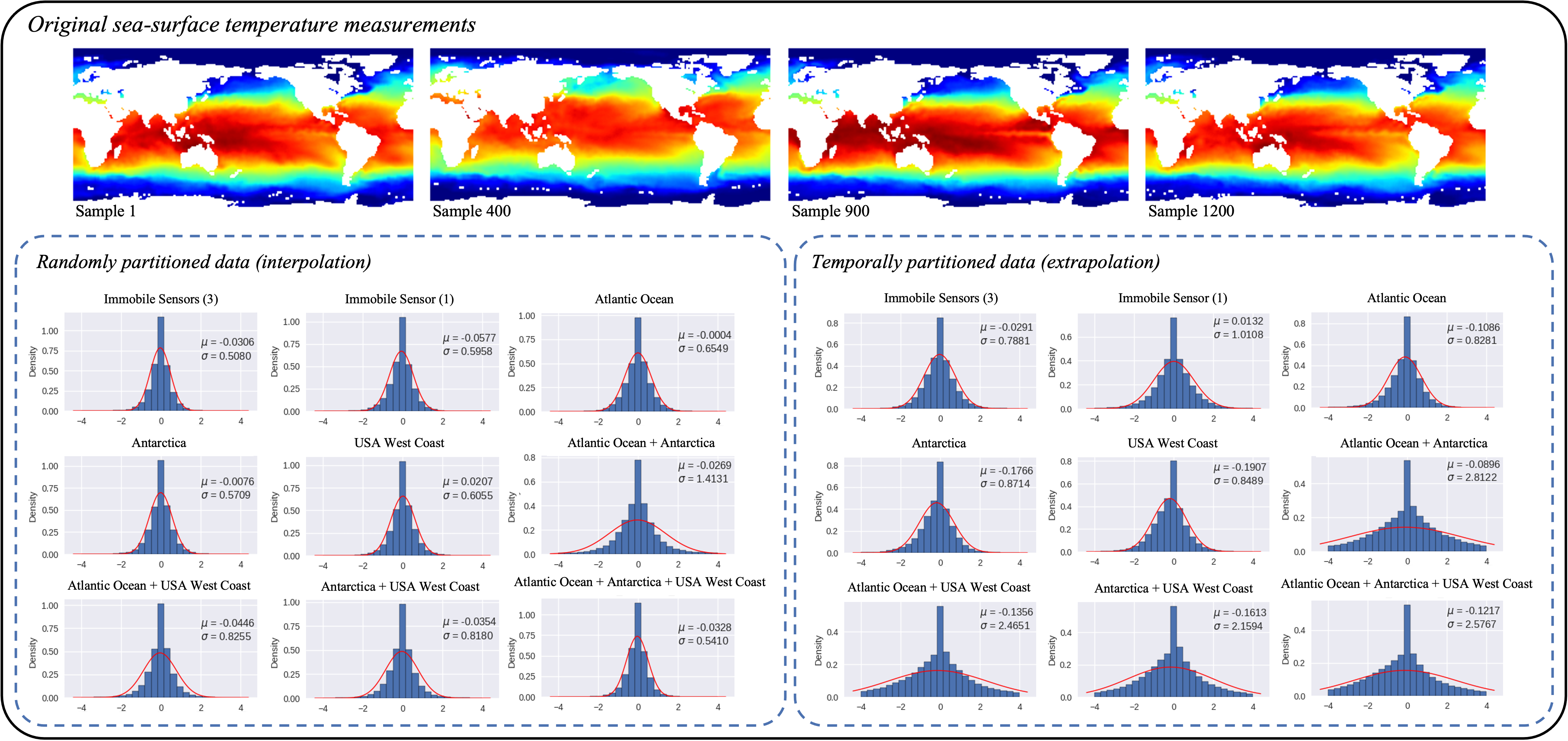}
    \caption{(Top) Example snapshots of global sea-surface temperature to be reconstructed from either the (i) randomly or (ii) temporally partitioned test set. (Bottom left) Histograms of the difference between ground truth and reconstructed states across all grid spaces and snapshots in the randomly-partitioned test set for immobile (1 and 3) and mobile sensors; dynamical trajectories of mobile sensors are one or more combinations of a year-long circuit in (i) the \textit{Atlantic Ocean}, \textit{Antarctica} Southern Ocean, and along the \textit{USA West Coast}. (Bottom right) Histograms of the differences between ground truth and reconstructed states across all grid spaces and snapshots in the temporally-partitioned test set, again for immobile and mobile sensors; sensor positions and trajectories were the same as the randomly-partitioned dataset.}
    \label{fig:SST}
\end{figure*}

\begin{table*}[]
\resizebox{0.8\textwidth}{!}{%
\begin{tabular}{|ccccc|}
\hline
\multicolumn{5}{|l|}{\textit{Marker-based Dataset} results reconstructing kinematic states} \\ \hline
\multicolumn{1}{|c|}{\multirow{2}{*}{Type}} &
  \multicolumn{1}{c|}{\multirow{2}{*}{Input Sensors}} &
  \multicolumn{3}{c|}{MSE ($\pm$ s.d.) of Rotational Kinematics [$^{\circ}$]} \\ \cline{3-5} 
\multicolumn{1}{|c|}{} &
  \multicolumn{1}{c|}{} &
  \multicolumn{1}{c|}{SHRED} &
  \multicolumn{1}{c|}{SDN} &
  Linear \\ \hline
\multicolumn{1}{|c|}{\multirow{4}{*}{Individual}} &
  \multicolumn{1}{c|}{\begin{tabular}[c]{@{}c@{}}3 random mobile sensors\\ (transverse-plane pelvis rotation angle, \\ medio-lateral pelvis position, \\ right hip adduction angle)\end{tabular}} &
  \multicolumn{1}{c|}{0.056 $\pm$ 0.018} &
  \multicolumn{1}{c|}{0.220 $\pm$ 0.097} &
  {0.419 $\pm$ 0.243} \\ \cline{2-5} 
\multicolumn{1}{|c|}{} &
  \multicolumn{1}{c|}{\begin{tabular}[c]{@{}c@{}}3 non-random mobile sensors\\ (right hip flexion angle, \\ right knee flexion angle, \\ right ankle dorsiflexion angle)\end{tabular}} &
  \multicolumn{1}{c|}{0.064 $\pm$ 0.027} &
  \multicolumn{1}{c|}{0.153 $\pm$ 0.097} &
  0.514 $\pm$ 0.328 \\ \cline{2-5} 
\multicolumn{1}{|c|}{} &
  \multicolumn{1}{c|}{\begin{tabular}[c]{@{}c@{}}1 random mobile sensor\\ (medio-lateral pelvis position)\end{tabular}} &
  \multicolumn{1}{c|}{0.080 $\pm$ 0.031} &
  \multicolumn{1}{c|}{1.033 $\pm$ 0.440} &
  1.096 $\pm$ 0.470 \\ \cline{2-5} 
\multicolumn{1}{|c|}{} &
  \multicolumn{1}{c|}{\begin{tabular}[c]{@{}c@{}}1 non-random mobile sensor\\ (right ankle dorsiflexion angle)\end{tabular}} &
  \multicolumn{1}{c|}{0.087 $\pm$ 0.042} &
  \multicolumn{1}{c|}{0.607 $\pm$ 0.244} &
  1.083 $\pm$ 0.705 \\ \hline
\multicolumn{1}{|c|}{\multirow{2}{*}{Population}} &
  \multicolumn{1}{c|}{\begin{tabular}[c]{@{}c@{}}3 non-random mobile sensors\\ (right hip flexion angle, \\ right knee flexion angle, \\ right ankle dorsiflexion angle)\end{tabular}} &
  \multicolumn{1}{c|}{0.098 $\pm$ 0.062} &
  \multicolumn{1}{c|}{0.311 $\pm$ 0.168} &
  0.944 $\pm$ 0.847 \\ \cline{2-5} 
\multicolumn{1}{|c|}{} &
  \multicolumn{1}{c|}{\begin{tabular}[c]{@{}c@{}}1 non-random mobile sensor\\ (right ankle dorsiflexion angle)\end{tabular}} &
  \multicolumn{1}{c|}{0.121 $\pm$ 0.073} &
  \multicolumn{1}{c|}{1.208 $\pm$ 0.822} &
  1.843 $\pm$ 2.110 \\ \hline
\end{tabular}%
}
\caption{Results from the human biomechanics dataset \cite{rosenberg2020predicting}. SHRED with random and non-random measurement trajectory inputs is compared to a SDN and linear model for reconstructing kinematic states for individual-specific models and population-based models. Overall, SHRED successfully reconstructed kinematic states (rotational variables showed here) with low mean-squared error, far outperforming the other architectures. Translational kinematic variables (x, y, and z-direction pelvis position) were excluded from these MSE calculations, as rotational and translational variables have different units ([$^{\circ}$] and [$m$], respectively; see Figs. \ref{fig:biomechanics_ind} and \ref{fig:biomechanics_pop} for all kinematic results.}
\label{tab:biomechanics}
\end{table*}

\subsection{Sea-surface temperature}
For our second dataset, we evaluate sea-surface temperature (SST) as reported by NOAA \cite{reynolds2002improved}. This dataset includes the weekly mean sea-surface temperature from the years 1992 to 2019. Notably, NOAA produced the SST data using optimum interpolation with in situ and satellite data. The data comprises 1400 snapshots of a 180 by 360 grid; the exclusion of landmass leaves 44,219 spatial locations corresponding to the sea surface. As in \cite{williams2023sensing}, we denote SHRED's input trajectory length to include 52 temporal measurements, resulting in one year of measurement time histories. The remaining 1348 snapshots are partitioned into training, test, and validation sets, comprised of 1000, 174, and 174 snapshots respectively. Regarding the dynamic trajectories of the mobile sensors, we consider three distinct moving sensors that complete a circuit over a year span: (i) an \textit{Atlantic Ocean} route that travels from the east coast of the USA towards Europe and back, (ii) an out-and-back route near \textit{Antarctica} Southern Ocean, and (iii) a route that follows the \textit{West Coast of the USA}. 

In Table \ref{tab:SST}, we compare how well each route -- as well as their combinations -- reconstruct SST, relative to one and three immobile sensors, for both random and temporal data partitions. For the random partition of training/test/validation data -- an interpolation problem -- we observe comparable mean-squared reconstruction errors across all mobile sensor routes relative to the immobile sensor reconstruction errors. For the temporal partition -- an extrapolation problem -- we see comparable errors for the single dynamic trajectory routes relative to the immobile sensors. The 2+ route combinations demonstrate much higher reconstruction errors; such results could be a potential indicator of overfitting, or even demonstrative of the challenges of extrapolating complex spatio-temporal data. 

We additionally evaluated the mean and variance in the mean-squared reconstruction errors across routes and data-partitions, as seen in Fig. \ref{fig:SST}. For the randomly-partitioned data of global sea-surface temperature, immobile sensors are more biased than mobile sensors ($\mu = -0.0306$ 3 immobile sensors, $\mu = -0.0577$ 1 immobile sensor; across all mobile sensor routes, minimum $\mu = -0.0004$, maximum $\mu = 0.0446$, absolute mean $\mu = 0.0241$). However, the variance of the mobile versus immobile sensors are more variable ($\sigma^2 = 0.5080^2$ 3 immobile sensors, $\sigma^2 = 0.5958^2$ 1 immobile sensor; across all mobile sensor routes, minimum $\sigma^2 = 0.5410$, maximum $\sigma^2 = 1.4131$, mean $\sigma^2 = 0.7756$). For the temporally-partitioned data, immobile sensors are less biased than mobile sensors ($\mu = -0.0291$ 3 immobile sensors, $\mu = -0.0132$ 1 immobile sensor; across all mobile sensor routes, minimum $\mu = -0.0896$, maximum $\mu = 0.1907$, absolute mean $\mu = 0.1406$). We see comparable variance for the single dynamic trajectory routes relative to the immobile sensors; the 2+ route combinations have much higher error variance; ($\sigma^2 = 0.7881^2$ 3 immobile sensors, $\sigma^2 = 1.0108^2$ 1 immobile sensor; across all mobile sensor routes, minimum $\sigma^2 = 0.8281$, maximum $\sigma^2 = 2.8122$, mean $\sigma^2 = 1.7959$).

\begin{figure*}[t]
    \centering
    \includegraphics[width=\textwidth]{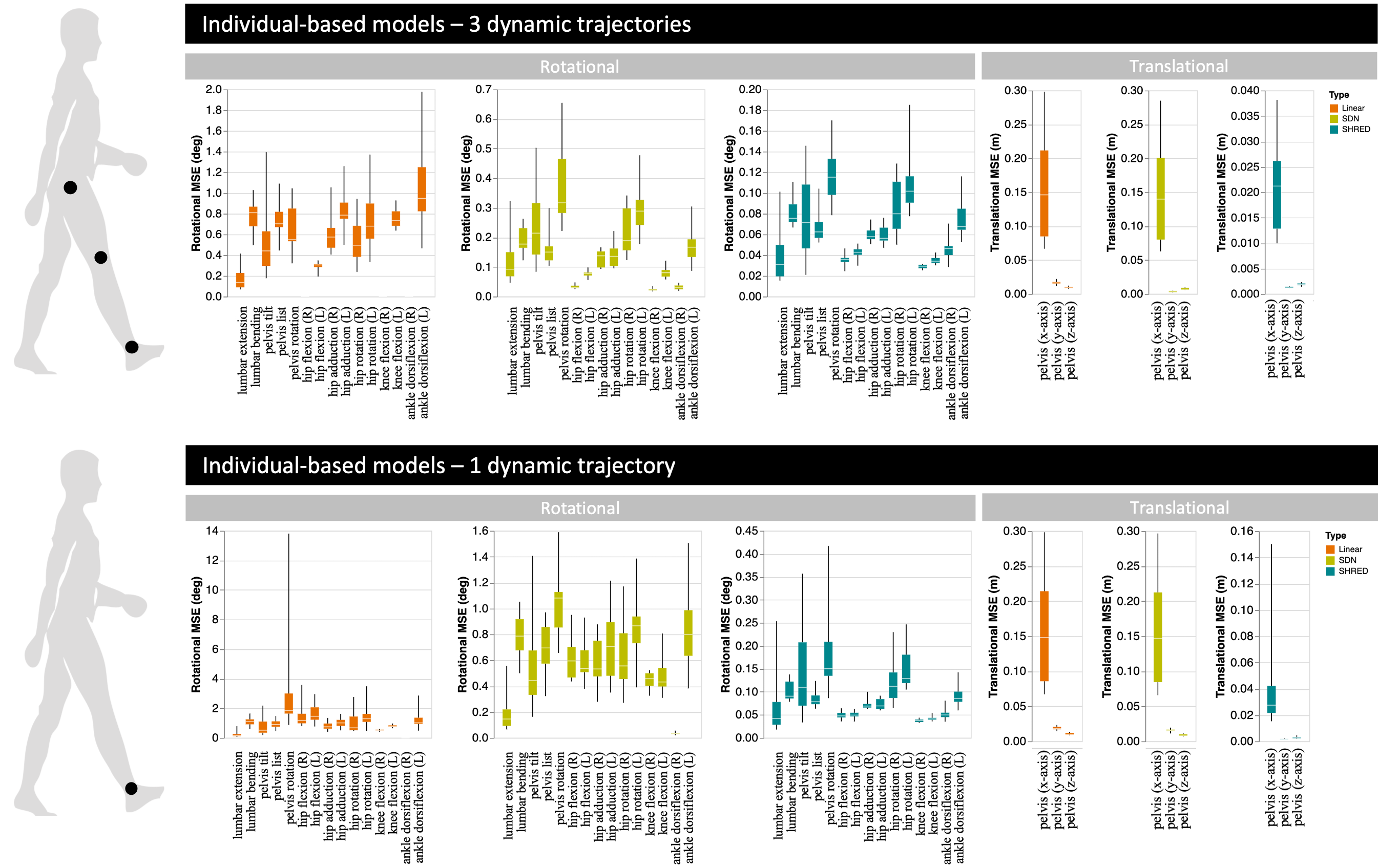}
    \caption{Visualization of individual-specific results with \textit{non-random} sensor inputs for reconstructing human biomechanics \cite{rosenberg2020predicting}; see Table \ref{tab:biomechanics} for additional individual-specific results using random sensor inputs. A unique mapping was trained for each individual to transform their sparse set of sensor measurements to their full measurement data; each individual's mapping was evaluated by reconstructing their held-out test data. Performance was calculated using mean-square error (MSE); box plots show the aggregate MSE for all individuals and across kinematic states. Rotational and translational kinematic states are displayed separately, as rotational states use units of degrees while translational states use units of meters. Three modeling paradigms were trained to learn a mapping: linear regression (orange), a shallow decoder network (SDN) (green), and a shallow recurrent demcoder network (SHRED) (blue). Note the independent y-axes and respective scales for each graph. (Top) Three dynamic trajectories were purposefully chosen as sensor inputs to the reconstruction mapping: right hip flexion angle (degrees), right knee flexion angle (degrees), and right ankle dorsiflexion angle (degrees). (Bottom) One dynamic trajectory was purposefully chosen as sensor inputs to the reconstruction mapping: right ankle dorsiflexion angle (degrees).}
    \label{fig:biomechanics_ind}
\end{figure*}

\begin{figure*}[t]
    \centering
    \includegraphics[width=\textwidth]{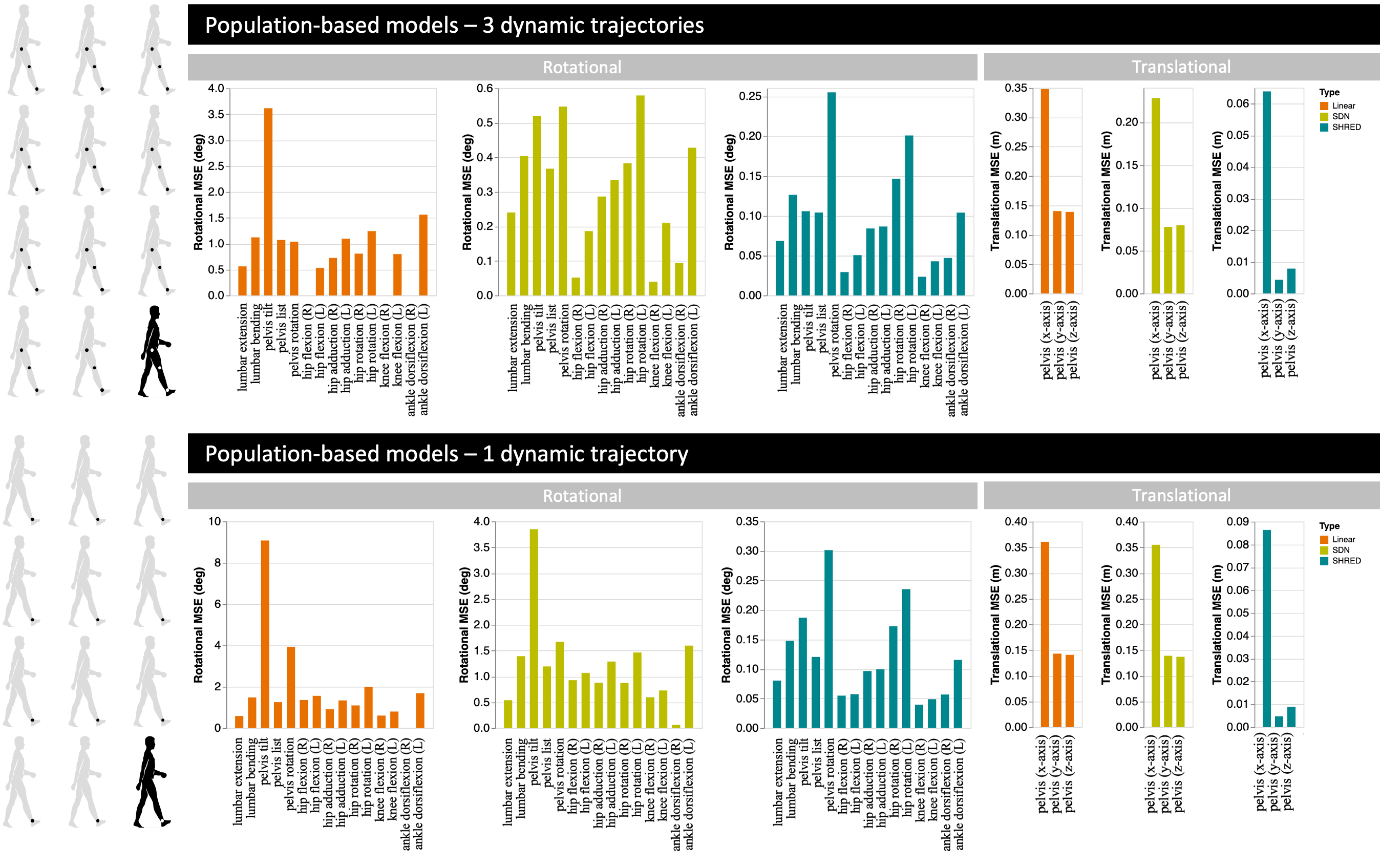}
    \caption{Visualization of population-based results for reconstructing human biomechanics \cite{rosenberg2020predicting}. The population models trained using 11 of the 12 individuals' data and were evaluated by reconstructing the held-out subject's full measurement data from their sparse input sensor(s); performance was calculated by using mean-square error (MSE). Rotational and translational kinematic states are displayed separately, as rotational states use units of degrees while translational states use units of meters. Three modeling paradigms were trained to learn the mapping: linear regression (orange), a shallow decoder neural network (green), and a shallow recurrent neural network (blue). Note the independent y-axes and respective scales for each graph. (Top) Three dynamic trajectories were purposefully chosen as sensor inputs to the reconstruction mapping: right hip flexion angle (degrees), right knee flexion angle (degrees), and right ankle dorsiflexion angle (degrees). (Bottom) One dynamic trajectory was purposefully chosen as sensor inputs to the reconstruction mapping: right ankle dorsiflexion angle (degrees).}
    \label{fig:biomechanics_pop}
\end{figure*}

\subsection{Human biomechanics}
The final dataset for which we leverage mobile sensor trajectories with SHRED is for capturing human biomechanics. Human motion tracking and analysis is essential for monitoring disease progression, guiding rehabilitation treatment, evaluating sports performance, and informing assistive device design. 
We use an open-source dataset that captures nondisabled human biomechanics during steady, rhythmic walking; such an approach can be generalized to other motion tracking such as robotic manipulation or computer animation. We use kinematics from 12 nondisabled adults (six female/six male; age 23.9 $\pm$ 1.8 years; height = 1.69 $\pm$ 0.10m; mass = 66.5 $\pm$ 11.7 kg) \cite{rosenberg2020predicting}. Participants walked at a self-selected speed (speed = 1.36 $\pm$ 0.11 m/s) on a split-belt instrumented treadmill (Bertec Corp., Columbus, USA) for six minutes. While kinematics can be captured with multiple modalities, gold-standard marker data was recorded using a 10-camera optical motion capture system (Qualisys AB, Gothenburg, SE). Kinematics were estimated from marker data using the Inverse Kinematics algorithm in OpenSim 3.3 with a dynamically-constrained 19 degree-of-freedom skeletal model scaled to each participant \cite{delp2007opensim, hicks2015my}. We evaluated the 18 kinematic states previously calculated in the open-source dataset which included three translational degrees of freedom at the pelvis, pelvis tilt/list/rotation, lumbar extension/bending, bilateral hip flexion/adduction/rotation, bilateral knee flexion, and bilateral ankle dorsiflexion. Kinematics were low-pass filtered at 6Hz using a fourth-order Butterworth filter \citep{rosenberg2020predicting}. An important distinction with this human biomechanics data, as opposed to the isotropic turbulence or sea-surface temperature, is that marker data recorded during motion capture is in a global coordinate frame, but through the standard procedures for marker data processing are transformed into kinematics in a relative frame of reference to the human subject.

Because this kinematic dataset cannot be compared to SHRED with immobile sensors, as the markers must move with the human, we compare the performance of mobile sensor trajectories and SHRED with that of a shallow decoder network (SDN) (\textit{i.e.}, SHRED without leveraging time histories) and a linear model. Therefore, the aim of this example was to evaluate each modeling architecture's ability to learn a mapping between a sparse set of mobile sensors to reconstruct the full set of 18 kinematic states. We evaluate two conditions: (1) individual-specific models, in which a model is developed for each participant to reconstruct their respective set of full kinematics, and (2) population-based models, which uses data from all participants to reconstruct the full set of kinematics for an unseen individual, thus enabling rapid generalization to data outside the training set. 

Overall, SHRED successfully maps between a sparse set of measurements to reconstruct the full biomechanical states for both the individual and population models. Additionally, SHRED far outperforms the linear and SDN architectures. While more input sensors generally increases reconstruction accuracy, performance with as little as \textcolor{black}{0.068 and 0.048} degrees (rotational variables) mean-squared error can be achieved with one and three sensors with SHRED, well below the accuracy and repeatability recommended even for clinical gait analyses ($2\deg$ in sagittal-plane and $5\deg$ in frontal-plane) \cite{schmitz2014accuracy, hicks2015my}. Table \ref{tab:biomechanics} provides a summary of the mean-squared errors for the rotational kinematic variables across modeling architectures and conditions. We take a closer look at each condition's results in the following sections:

\subsubsection{Individual-specific biomechanics models}

For the individual-specific biomechanics models, a mapping is trained for each of the 12 participants using a linear, SDN and SHRED architecture. Four combinations of dynamic trajectory inputs are tested: (1) three randomly chosen kinematic states (transverse-plane pelvis rotation angle, medio-lateral pelvis position, right hip adduction angle), (2) three non-randomly chosen kinematic states (right hip flexion angle, right knee flexion angle, right ankle dorsiflexion angle), (3) one randomly chosen kinematic state (medio-lateral pelvis position), and (4) one non-randomly chosen kinematic state (right ankle dorsiflexion angle). The non-randomly chosen states represent the kinematics most commonly used for biomechanical assessments, and which generally have the greatest accuracy and inter-session reliability (i.e, sagittal-plane kinematics) \cite{schmitz2014accuracy}. 

Fig. \ref{fig:biomechanics_ind} shows the performance of SHRED relative to the SDN and linear model for reconstructing human motion across individuals. SHRED reconstructs the full set of 18 biomechanical variables more accurately that the shallow decoder and linear models, regardless of the number or choice of input sensors. Using three mobile sensors, SHRED has errors less than $0.064 \pm 0.027$ degrees (rotational variables), as much as $4.0$x more accurate on average than the SDN and $8.0$x more accurate than the linear model. Using a single mobile sensor, SHRED has errors less than $0.087 \pm 0.042$ degrees, as much as $12.8$x more accurate on average than the SDN and $13.6$x more accurate than the linear model.

\subsubsection{Population-based biomechanics models}

For the population-based biomechanics models, a reconstruction mapping was trained using 11 of the 12 participants' kinematic data, for a one-subject hold-out test using each architecture. Two combinations of dynamic trajectory inputs were tested: (1) three non-randomly chosen kinematic states (right hip flexion angle, right knee flexion angle, right ankle dorsiflexion angle), and (2) one non-randomly chosen kinematic state (right ankle dorsiflexion angle). 

Fig. \ref{fig:biomechanics_pop} shows the performance of SHRED relative to the SDN and linear model for reconstructing human motion within a population. SHRED far outperforms the SDN and linear model regardless of number or choice of input sensors. Using three mobile sensors, SHRED has errors less than $0.098 \pm 0.062$ degrees, which was on average $3.3$x more accurate than the SDN and $9.6$x more accurate than the linear model. Using a single mobile sensor, SHRED has errors less than $0.121 \pm 0.073$ degrees, which was on average $9.8$x more accurate than the SDN and $14.8$x more accurate than the linear model. These results demonstrate that SHRED can enable rapid generalization (parameterization of dynamics) for data outside the training set, which may be especially useful in human biomechanics where population-level models can be used to estimate parameters for a new individual (\textit{e.g.}, a new patient visits a clinic for the first time).

\section{Discussion \& Conclusion}

Sensing continues to be one of the most important tasks for science and engineering across disciplines.  Machine learning methods are offering exceptional new paradigms for maximally exploiting sensor information for the diversity of tasks required of sensors, including reconstruction, forecasting, model discovery, control, and uncertainty quantification.  Although there is a well established theory for sensing and sensor placement using immobile sensors, mobile sensing has remained exceptionally challenging with only recent and limited advancements made for leveraging the dynamical trajectories of the sensor.  The mobile SHRED architecture advocated here provides a new paradigm for sensing whereby the time-history of the sensor is used to encode global information of the measured high-dimensional state space.   SHRED not only allows for the embedding of the multiscale physics into a compact and low-dimensional latent space, it also provides through its decoder network a mapping from the minimal mobile sensors to full state estimates.  Thus the SHRED architecture provides a nonlinear generalization of classical low-rank and linear embeddings such as the singular value decomposition.  

The performance of SHRED is demonstrated on three challenging data sets which are all characterized by complex, nonlinear, and multiscale interactions.  Thus the data considered represents some of the most challenging environments for sensing. Additionally, such applications often require the mobile sensing.  Given the limited work on mobile sensing to date, along with the performance demonstrated by the method, the mobile SHRED architecture represents a significant innovation in mobile sensing technologies.  Not only does the architecture reduce the variance in state estimation in comparison to immobile sensors, the decoder requires less data to train than typical networks due to its shallow structure.  Empirically it is also observed that the mobile SHRED architecture is robust to hyper-parameter tuning, requiring little effort to train high performing models.  The advantageous of such an architecture provide an application agnostic scheme which has significant potential for broad usage in science and engineering.

The mobile SHRED architecture works because the time history of the sensors embed global information of the system measured.  For spatio-temporal data, for instance, the dynamics typically are governed by partial differential equations whose spatial derivative estimates rely on neighboring data points.  These neighbors in turn rely on their neighbors and so forth.  Thus a local spatial position is coupled to neighboring positions and ultimately globally by spatial derivatives.  The sensor trajectories learn a representation of the spatio-temporal field via this global coupling.  Of course, such an architecture would break down if there are regions in the dynamics that are statistically independent from each other, \textit{i.e.} there are isolated dynamics that are not coupled in any way. There can also be ``dead spots" in the measurement space, but the moving sensors can then typically move to new locations where information is captured.  Thus with high-probability, a random sensor trajectory can be used to encode global information and estimate the full state space.

Finally, for the examples shown, there is clearly a requirement in the training data to have both the sensor trajectory data as well as the corresponding high-dimensional state space data.  In many applications, there is no access to the high-dimensional state space.  Thus proxy data, in the form of simulation data, can be used as a surrogate for the real data.  As long as the proxy data is representative of the true data, it can be used to train the SHRED architecture.  This gives a hybrid approach to training the neural network whereby simulation data can be used in partnership with real data to build a sensing model for producing otherwise inaccessible full state estimates.

\section*{Acknowledgements}

JNK acknowledges support from the Air Force Office of Scientific Research (FA9550-19-1-0386).
The authors also acknowledge support in part by the US National Science Foundation (NSF) AI Institute for Dynamical Systems (dynamicsai.org), grant 2112085.

\bibliography{references}

\end{document}